\pdfoutput=1
\documentclass{svmult}
\usepackage{graphicx}
\usepackage{amsmath}
\usepackage{amssymb}

\newcommand{\ra}{\rangle}
\newcommand{\la}{\langle}
\newcommand{\ttt}{\texttt}

\begin{document}

\title*{A Reflection on the Structure and Process of the Web of Data}
\author{Marko A. Rodriguez}

\institute{
	T-5, Center for Nonlinear Studies \\ 
	Los Alamos National Laboratory \\
	Los Alamos, New Mexico 87545 \\
	\texttt{marko@lanl.gov}\\
}

\maketitle

\footnotetext[1]{Rodriguez, M.A., ``A Reflection on the Structure and Process of the Web of Data," Bulletin of the American Society for Information Science and Technology, volume 35, number 6, pages 38--43, ISSN:1550-8366, LA-UR-09-03724, September 2009.}

The Web community has introduced a set of standards and technologies for representing, querying, and manipulating a globally distributed data structure known as the Web of Data. The proponents of the Web of Data envision much of the world's data being interrelated and openly accessible to the general public. This vision is analogous in many ways to the Web of Documents of common knowledge, but instead of making documents and media openly accessible, the focus is on making data openly accessible. In providing data for public use, there has been a stimulated interest in a movement dubbed Open Data \cite{opendata:miller2008}. Open Data is analogous in many ways to the Open Source movement. However, instead of focusing on software, Open Data is focused on the legal and licensing issues around publicly exposed data. Together, various technological and legal tools are laying the groundwork for the future of global-scale data management on the Web. As of today, in its early form, the Web of Data hosts a variety of data sets that include encyclopedic facts, drug and protein data, metadata on music, books and scholarly articles, social network representations, geospatial information, and many other types of information. The size and diversity of the Web of Data is a demonstration of the flexibility of the underlying standards and the overall feasibility of the project as a whole. The purpose of this article is to provide a  review of the technological underpinnings of the Web of Data as well as some of the hurdles that need to be overcome if the Web of Data is to emerge as the \textit{defacto} medium for data representation, distribution, and ultimately, processing.

Technically, on the Web of Data, Uniform Resource Identifiers (URI) are used to identify resources \cite{uri:berners2005}. For example, depending on what is being modeled, a URI can denote a city, a protein, a music album, a scholarly article, a person, etc. In general, any thing can be assigned a URI. An example URI is \ttt{http://www.lanl.gov\#marko}. This URI denotes the author of this article, Marko. The URI has information pertaining to the what (\ttt{marko}), where (\ttt{www.lanl.gov}), and how (\ttt{http}) of a resource. The URI is more general than the URL of common knowledge as URIs are not required to resolve to retrievable digital objects such as documents and media. Instead, URIs can denote abstract concepts such as the person Marko, the class of dogs, or the notion of friendship. Finally, the space of all URIs is an inherently distributed and theoretically infinite space. This makes the URI space fit to represent massive amounts of data distributed world wide. A convenient consequence of this space is that the Web of Data can emerge atop it. However, while URIs can denote things, they can not denote how things relates to each other. Relating URIs is necessary in order to give greater meaning and context to datum. Moreover, relating URIs is necessary to create the ``web" aspect of the Web of Data.

The Resource Description Framework (RDF) is a standardized data model for linking URIs in order to create a network/graph of space of all URIs \cite{rdfintro:miller1998}. RDF also supports the linking of URIs to primitive literals such as strings, integers, floating point values, etc. An example RDF statement to denote the fact that ``Marko knows Fluffy" is $\la$\ttt{http://www.lanl.gov\#marko}, \ttt{http://xmlns.com/foaf/0.1/knows}, \ttt{http://www.lanl.gov\#fluffy}$\ra$. In order to make long URIs more readable, namespace prefixes are generally used. With namespace prefixes, the previous statement can be represented as $\la \ttt{lanl:marko}, \ttt{foaf:knows}, \ttt{lanl:fluffy} \ra$. All RDF statements have this three component form, where there exists a subject (\ttt{lanl:marko}), a predicate (\ttt{foaf:knows}), and an object (\ttt{lanl:fluffy}). As such, an RDF statement is also known as a triple. A URI can be involved in multiple statements. For example, it is possible to state that, while being known by Marko, Fluffy is also a dog, 5 years old, and lives in Santa Fe, New Mexico. Data on Santa Fe and data on Fluffy become merged when statements involving their two URIs are joined (directly or indirectly through multiple links). The Web of Data becomes powerful when seemingly different data sets are interlinked. The fact that Fluffy lives in Santa Fe automatically connects data about Fluffy to geographic and encyclopedic data about Santa Fe, New Mexico---its geospatial coordinates, nearby cities, culture, population, etc. As more and more statements are added to the Web of Data, the Web of Data serves, in a sense, as a global database of interlinked heterogeneous data. The combination of both the URI and RDF has moved the World Wide Web beyond a Web of Documents to that of a Web of Data, where every minutia of information can be represented and interlinked for consumption by both man and machine.
 
RDF's original use case has evolved beyond that of a logic-language for knowledge representation and reasoning on the Semantic Web \cite{webinterpret:rodriguez2009}. As the foundational technology of the Web of Data, RDF can be seen as a general-purpose data model. It can be used to model formal knowledge (the Semantic Web), graph/networks (the Giant Global Graph), and software and abstract virtual machines (the Web of Programs) to name a few. In many ways, URIs and RDF afford a memory structure analogous to the local memory of a physical machine except that this memory structure is distributed over physical machines world wide. Each physical machine stores and manages a subset representation of the full Web of Data. RDF can be stored on a physical machine in many ways. A simple, straightforward way is to represent RDF statements in a file---an RDF document. A common misconception is that RDF and RDF/XML are one in the same. RDF is a data model that has various serialized representations with RDF/XML being one such serialization. Other popular serializations include N3 and N-Triple. Thus, there are many types of RDF documents. For the small-scale exposure of RDF data, an RDF document suffices. For the large-scale exposure and processing of RDF data, an RDF repository known as a triple store or graph database is usually the chosen solution. The expanded use of RDF has been greatly facilitated by the continued increase in the capacity and speed of RDF triple stores. Modern high-end RDF triple stores can hold and process on the order of 10 billion triples. Example high-end triple stores include Neo4j\footnote{Neo4j is available at \ttt{http://neo4j.org/}.} and AllegroGraph\footnote{AllegroGraph is available at \ttt{http://www.franz.com/agraph/allegrograph/}.}. What has been the sole territory of relational database technologies may soon be displaced by the use of RDF and the triple store. Moreover, because RDF is the common data model utilized by triple stores, it is possible to integrate data sets across different triple stores---across different RDF data providers. This integration is conveniently afforded by the URI and RDF as Web standards and is a function foreign to the relational database domain. With the Web of Data, no longer is information isolated in individual inaccessible data silos, but is instead exposed in an open and interconnected environment---the Web environment. The means to integrate RDF data across different RDF data sets is explained next.

\section{Linked Data and a Distributed Data Structure}

\begin{figure*}[h!]
	\centering
		\includegraphics[width=1.0\textwidth]{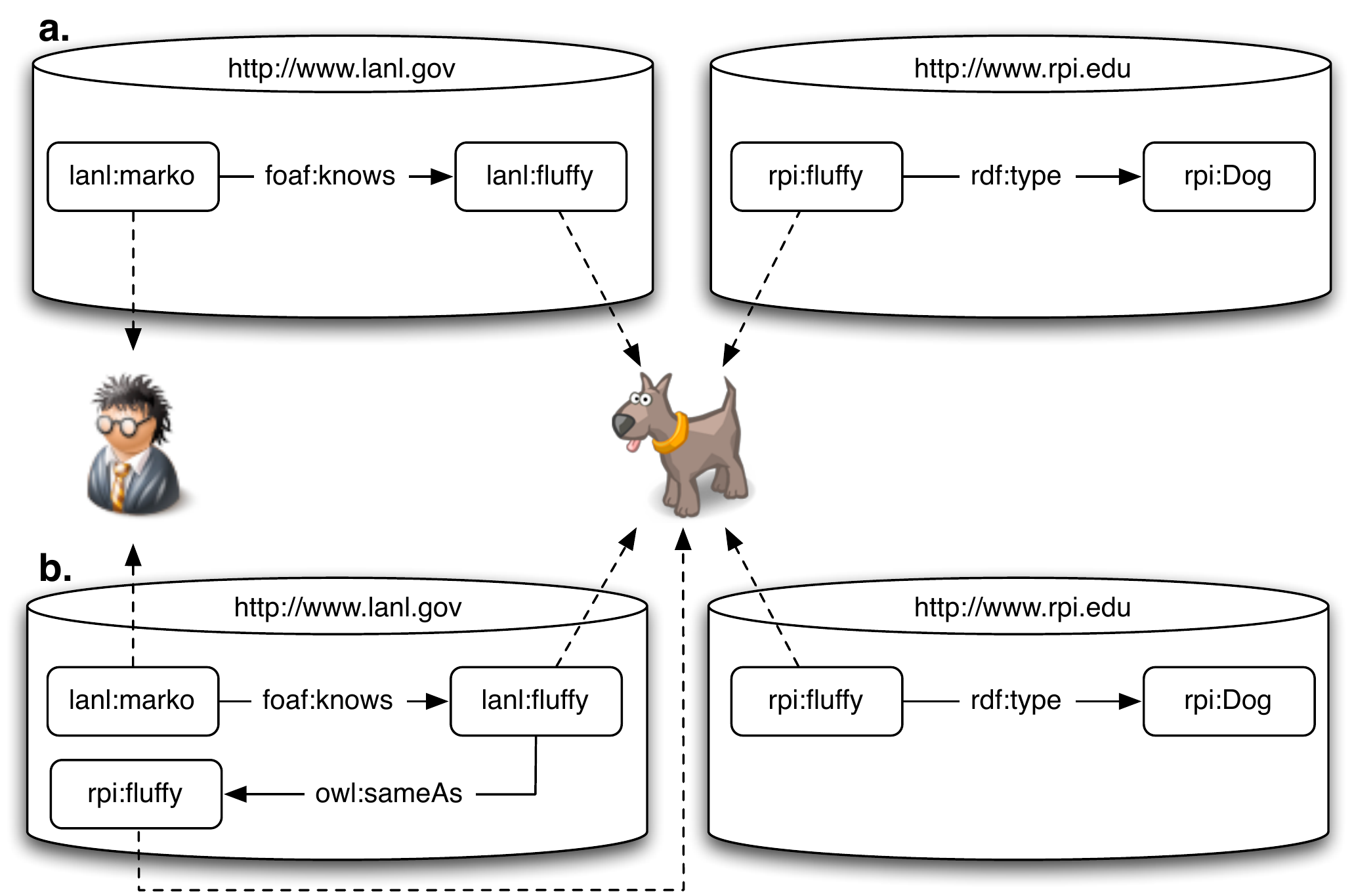}
	\caption{Two RDF repositories utilize different URIs to denote the same thing. By making it explicit that two URIs denote the same thing (i.e. $\la \ttt{lanl:fluffy}, \ttt{owl:sameAs}, \ttt{rpi:fluffy} \ra$), it is possible to merge data sets together. This merging of data sets is what creates the Web of Data.\label{fig:uri-denotes}}
\end{figure*}

In an effort to provide a seamless integration between the data provided by different RDF data providers, the Linked Data community is focused on developing the specifications and tools for linking RDF data sets into a single, global ``web of data" \cite{linkeddata:bizer2008}. Two RDF data sets link together when one data set uses a URI maintained by another. For example, suppose the URI \ttt{lanl:fluffy} minted and maintained by the Los Alamos National Laboratory (LANL). As previously explained, this URI is denoting something in the world---namely Fluffy. However, it is possible for someone other than LANL to express statements about Fluffy. Assume that the Rensselaer Polytechnic Institute (RPI) mints their own URI \ttt{rpi:fluffy} to denote Fluffy, where \ttt{rpi} is the namespace prefix that resolves to \ttt{http://www.rpi.edu\#}. At this point, both \ttt{lanl:fluffy} and \ttt{rpi:fluffy} denote the same thing---they both denote the same real-world object known as Fluffy. This idea is diagrammed in Figure \ref{fig:uri-denotes}a, where the dashed lines identify which worldly things the URIs stand in reference to. In order to link the LANL data set with the RPI data set, LANL can add the RDF statement $\la \ttt{lanl:fluffy}, \ttt{owl:sameAs}, \ttt{rpi:fluffy}\ra$ to its data set. This statement states that both \ttt{lanl:fluffy} and \ttt{rpi:fluffy} denote the same real-world thing, Fluffy. This idea is diagrammed in Figure \ref{fig:uri-denotes}b. Given this statement, it is possible to traverse from \ttt{lanl:fluffy} (LANL) to \ttt{rpi:fluffy} (RPI) and thus, migrate from the LANL data set to the RPI data set. When two data sets denote the same thing, they can be linked.

The Linked Data community is interested in both unifying RDF data sets as well as specifying the behaviors associated with URI resolution. A Linked Data-compliant data provider should return data when a URI is dereferenced---when a representation of the resource being identified by the URI is requested. More specifically, when a URI is dereferenced, a collection of statements associated with that URI should be returned in some RDF serialization such as RDF/XML. Given the example above, if a machine dereferences \ttt{lanl:fluffy}, it will get the statement $\la$\ttt{lanl:fluffy}, \ttt{owl:sameAs}, \ttt{rpi:fluffy}$\ra$ returned to it. In other words, LANL returns all the RDF statements for which \ttt{lanl:fluffy} is the subject of the triple (i.e. the outgoing edges from \ttt{lanl:fluffy}). Now, the machine knows that \ttt{lanl:fluffy} and \ttt{rpi:fluffy} denote the same thing. Thus, if it wants to know what RPI has stated about Fluffy, it will dereference \ttt{rpi:fluffy}. Upon doing so, it should get the statement $\la$\ttt{rpi:fluffy}, \ttt{rdf:type}, \ttt{rpi:Dog}$\ra$ returned to it. To the machine, the Web of Data is one expansive interlinked web of URIs. To the underlying servers, the Web of Data is broken up into multiple RDF subgraphs (multiple data providers) and linked together when one data provider references a URI minted and maintained by another data provider. It is noted that resolving a Linked Data-compliant store's URI is one way of getting data from the Web of Data. For more complicated data gathering situations, many RDF data providers expose SPARQL end-points to their triple stores. SPARQL is a query language similar to SQL, but focused on graph queries as opposed to table queries \cite{sparql:prud2004}. 

\begin{figure*}[h!]
	\centering
		\includegraphics[width=1.0\textwidth]{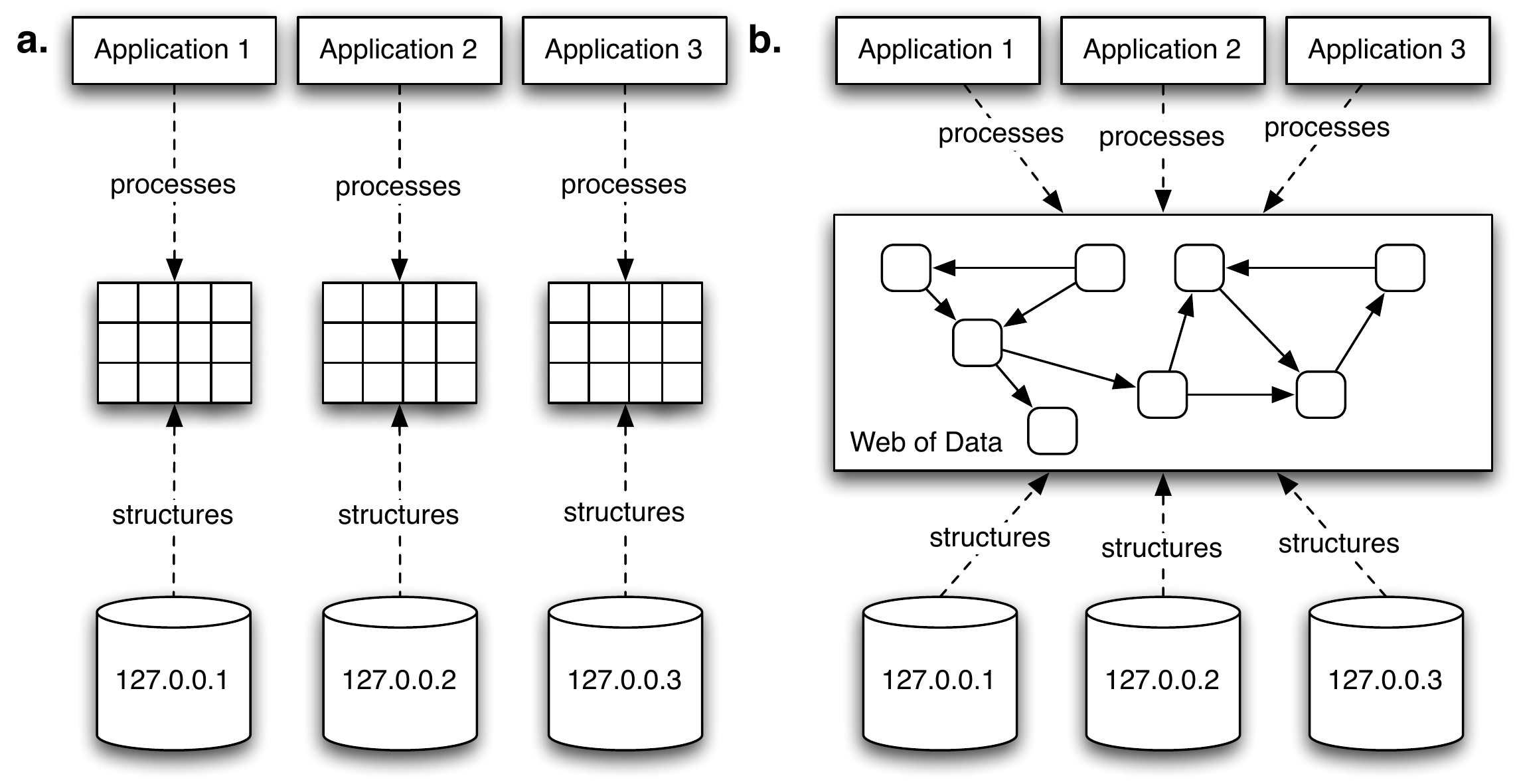}
	\caption{a.) The typical web application requires its own data source on which to provide its service. b.) On the Web of Data, application providers are cleanly separated from the data providers.\label{fig:structures-processes}}
\end{figure*}

An interesting consequence of the Web of Data is that it can greatly shift the role of application and data providers. Currently, web applications are required to maintain their own data source. For example, Amazon.com maintains its database of books, Springer its database of journal articles, and iTunes its database of music metadata. In order for users to utilize this data in interesting ways, these same data providers must provide a front-end application to interact with the data. In this way, data providers and application developers are one in the same entity. This idea is diagrammed in Figure \ref{fig:structures-processes}a, where each application utilizes its own back-end database to provide its front-end application with data. With the Web of Data, this model is significantly altered. On the Web of Data, application providers and data providers are cleanly separated. Data providers can provide and interlink book, article, and music data on the Web of Data and application providers can develop software to utilize this data for different end-user services---book recommendations, citation analysis, and music metadata population. Moreover, this same data can be utilized by multiple different application developers and thus, this can yield many ways for the end-user to interact with the Web of Data. In other words, Amazon.com's data may be more efficiently presented and processed if it was open for any developer to create a front-end application for it. This idea is diagrammed in Figure \ref{fig:structures-processes}b. The clean separation between data and application providers is already taking place as plenty of interlinked heterogeneous data currently exists on the Web of Data. A few examples are provided here. Book data can be found at Amazon.com's RDF BookMashup and the RDF representation of Project Guttenburg. Scholarly data is provided by the Digital Bibliography and Library Project (DBLP), ACM, IEEE, amongst many others. Finally, various music data sets exist such as MusicBrainz and AudioScrobbler. This data is leveraged, as mentioned previously, by resolving URIs. For example, if one were to dereference this URI in a standard web browser \ttt{http://rdf.freebase.com/ns/guid.9202a8c04000641f800000000001a49d}, what is returned is a set of RDF statements (as an RDF/XML document) linking this URI to other URIs and literals. Accessible, interlinked, structured data is the point of the Web of Data. An ecology of applications leveraging this data may greatly advance applications and algorithms for processing data as no longer are application developers burdened by the ``cold start" problem of requiring large-amounts of data to initiate a successful service \cite{faith2:rodriguez2009}. No longer will consumers be confined to use certain web applications as no longer are applications and data so tightly coupled.

\section{Linked Process and a Distributed Process Infrastructure}

While the Web of Data and the efforts expended by the Linked Data community have provided a path towards global-scale data management, this model is lacking one important component: an infrastructure for data processing. A significant hurdle to overcome for this community is that of distributed processing on this distributed data structure. Traversing the Web of Data is not quite the same as traversing the Web of Documents. For the human, it is reasonable to traverse from URI to URI exploring the Web of Data in a manner similar to how the Web of Documents is traversed. That is, a human, using their Web browser, can resolve URIs and view the RDF data returned. Moreover, various human-friendly RDF browsers exist (usually in the form of a Web browser plugin) to make it easy for humans to view and traverse the data on the Web of Data. However, for a machine (i.e. an application, an algorithm), the Web of Data can be traversed much faster than what a human can do by manually clicking from URI to URI. Moreover, there will be orders of magnitude more resources and links on the Web of Data than what is found on the Web of Documents. While a machine can crawl and pull the data to its local environment for processing, this becomes inefficient when the data requirements span large parts of the Web of Data. Again, note that every time a URI is dereferenced, the resolving server prepares an RDF subgraph and returns it (over the wire) to the requesting machine. Thus, ``traversing" the Web of Data requires data to be migrated to the traversing machine and processed remotely from the data source. This architecture is analogous to the current Web of Documents whereby ``traversing" the Web of Documents pulls HTML documents and media to the requesting machine. For human consumption, this is necessary as data/documents must be rendered where the human is physically located---remote from the data source. For a machine (a virtual machine) its physical location need not be a factor in how data is consumed and processed. Thus, for processing large parts of a distributed data structure, a more efficient mechanism would be to migrate the process between data providers so that information is not pulled over the wire, but instead, processed where the data is maintained. In other words, an efficient mechanism for processing the Web of Data would be to move the process to the data, not the data to the process \cite{rodriguez:gpsemnet2009}.

For the Web of Documents, the search engine philosophy of ``download and index" has made it possible for end users to find information in a more efficient manner than by simply surfing and bookmarking. With modern commercial triple stores scaling to the order of 10 billion triples, centralized indexing repositories will have to contend with not only the volume of data, but also the computational complexities of analyzing such data in sophisticated ways. The Web of Data provides a much richer machine processable data structure than what is provided by HTML and thus, antiquated keyword search simply does not take significant advantage of what the Web of Data is providing. The future of the Web of Data will be rife with algorithms from many schools of thought---formal logic, graph analysis, object-oriented programming, etc. \cite{webinterpret:rodriguez2009}. Many of these algorithms will compute across various underlying stores of the Web of Data and will require a distributed Turing complete infrastructure to do so. For any algorithm of sufficient complexity, there is simply too much data to pull over the wire and thus, the Web of Data in its current form greatly reduces what is possible. This is an unfortunate state of affairs. Given the potential role of the Web of Data as the \textit{defacto} medium for interconnecting data, a distributed computing environment is necessary. The Linked Data community needs a parallel Linked Process effort. In a sense, data providers already expose their processors for public use by way of their SPARQL-endpoints. SPARQL serves as an on-site data processing language. However, this language, being a query language, is not sufficient for representing complex algorithms. What is needed is a framework that is more general-purpose and respective of the three following basic requirements:
\begin{enumerate}
	\item safe: applications must not be able to destroy the integrity of the open processor or its data set when using this infrastructure.
	\item efficient: applications must run faster in this infrastructure than what is possible when pulling the required data over the wire.
	\item easy to use: application developers must be be able to utilize common programming languages and packages and be relatively blind to the underlying infrastructure.
\end{enumerate}

Developing a distributed process infrastructure that accounts for these three factors will ultimately drive its adoption. With the widespread adoption of such a processing infrastructure by RDF data providers, the Web of Data will reach a new level of functionality. No longer will the Web of Data be only a database serving data over the wire to third-party applications, but instead, a distributed computing environment supporting complex algorithms that can leverage rich data in ways not previously possible in the history of computing. The unification of Linked Data and Linked Process in many ways is similar to ``cloud computing." However, with the integration of data sets and hardware processors world wide, this ``cloud" will be much richer and more decentralized than what exists with other cloud providers. In this form, the Web of Data will afford the world a democratization of both data and process and may perhaps enjoy a frenzied adoption similar to what has occurred with its predecessor, the Web of Documents.

\section{Conclusion}

The Web of Data provides an infrastructure that supports an instantiation of a distributed graph of Web resources. This distributed graph is created by many data providers representing and interrelating their data. What emerges from this collective effort is a publicly accessible global database that can be leverage by both man and machine to any end they deem appropriate. However, the current instantiation of the Web of Data lacks one crucial component: a distributed processing infrastructure. For the Web of Documents of common knowledge, the solution to the issue of processing the vast amount of information has been to literally download the entire Web and index and process it in a single environment. While the content on the Web of Documents is distributed, the means by which the information on the Web of Documents is analyzed is not. The Web of Data need not fall into this same model. With the nearly limitless ways in which RDF data can be processed, it would be a disappointment if the data on the Web of Data was left solely to centralized repositories to store, index, and provide query functionality. Beyond disappointment, it would reduce the potential utility the Web of Data would have given a distributed process infrastructure. By extending the work of the Linked Data community with Linked Process, the Web of Data may one day rise to become the \textit{defacto} medium for representing and processing data much like the Web of Documents is the \textit{defacto} medium for storing and sharing documents.

\end{document}